\documentclass[11pt,a4paper]{article}
\usepackage[hyperref]{acl2019}
\usepackage[utf8]{inputenc}
\usepackage{times}
\usepackage{latexsym}
\usepackage{amsmath}
\usepackage{graphicx}
\usepackage{amssymb}
\usepackage{comment}

\usepackage{url}

\aclfinalcopy 

 \title{Empirical Study of Diachronic Word Embeddings for Scarce Data}

\author{Syrielle Montariol \\
  LIMSI - CNRS \\
  Univ. Paris-Sud, Univ. Paris-Saclay \\
  Société Générale \\
  {\tt syrielle.montariol@limsi.fr} \\\And
  Alexandre Allauzen \\
  LIMSI - CNRS \\
  Univ. Paris-Sud, Univ. Paris-Saclay \\
  {\tt alexandre.allauzen@limsi.fr} \\}

\date{}

\begin{document}
\maketitle

\begin{abstract}
Word meaning change can be inferred from drifts of time-varying word embeddings.
However, temporal data may be too sparse to build robust word embeddings and to discriminate significant drifts from noise. In this paper, we compare three models to learn diachronic word embeddings on scarce data: incremental updating of a Skip-Gram from \newcite{kim2014}, dynamic filtering from \newcite{bamler17a}, and dynamic Bernoulli embeddings from \newcite{rudolph2018dynamic}. In particular, we study the performance of different initialisation schemes and emphasise what characteristics of each model are more suitable to data scarcity, relying on the distribution of detected drifts. Finally,  we regularise the loss of these models to better adapt to scarce data.
\end{abstract}

\section{Introduction}
In all languages, word usage can evolve over time, mirroring cultural or technological evolution of society~\cite{Aitchison}.

For example, the word "Katrina" used to be exclusively a first name until year 2005 when hurricane Katrina devastated the United States coasts. After that tragedy, this word started to be associated with the vocabulary of natural disasters.

In linguistics, \textit{diachrony} refers to the study of temporal variations in the use and meaning of a word. Detecting and understanding these changes can be useful for linguistic research, but also for many tasks of Natural Language Processing (NLP). Nowadays, a growing number of historical textual data is digitised and made publicly available. It can be analysed in parallel with contemporary documents, for tasks ranging from text classification to information retrieval. However, the use of conventional word embeddings methods have the drawback to average in one vector the different word's usages observed across the whole corpus. This \textit{static} representation hypothesis turns out to be limited in the case of temporal datasets.


Assuming that a change in the context of a word mirrors a change in its  meaning or usage, a solution is to explore diachronic word embeddings: word vectors varying through time, following the changes in the global context of the word. While many authors proposed diachronic embedding models these last years, these methods usually need large amounts of data to ensure robustness. 

However, temporal datasets often face the problem of scarcity; beyond the usual scarcity problem of domain-specific corpora or low-resource languages, a temporal dataset can have too short time steps compared to the volume of the full dataset.\footnote{A short time step can be one month or less, depending on the domain.} Moreover the amount of digital historical texts is limited for many languages, particularly for oldest time periods.

This paper addresses the following question: In case of scarce data, how to efficiently learn time-varying word embeddings?
For this purpose, we compare three diachronic methods on several sizes of datasets. The first method is incremental updating \cite{kim2014}, where word vectors of one time step are initialised using the vectors of the previous time step. The second one is the dynamic filtering algorithm \cite{bamler17a} where the evolution of the embeddings from one time step to another is controlled using a Gaussian diffusion process. Finally, we experiment dynamic Bernoulli embeddings \cite{rudolph2018dynamic} where the vectors are jointly trained on all time slices. 

These three models are briefly described in  section~\ref{sec:model}. The hyper-parameters are specifically tuned towards efficiency on small datasets. Then, we explore the impact of different initialisation scheme and compare the behaviour of word drifts exhibited by the models. Finally,  we experiment regularising the models in order to tackle the faults detected in the previous analysis.
The experiments in section ~\ref{sec:expe} are made on the \textit{New York Times Annotated Corpus} (NYT) \footnote{https://catalog.ldc.upenn.edu/LDC2008T19}  \cite{nytLDC} covering two decades. 

\section{Related Work}

The first methods to measure semantic evolution rely on detecting changes in word co-occurrences, and approaches based on distributional similarity \cite{Gulordava}.
The use of automated learning methods, based on word embeddings \cite{mikolovW2V}, is recent and has undergone an increase in interest these last two years with the successive publication of three articles dedicated to a literature review of the domain \cite{SOTAKutuzov, Tahmasebi2018SurveyOC, tang_2018}.
In this section, we mainly consider this second line of work, along with the peculiarities of scarce data. 

\newcite{kim2014} developed one of the first method to learn time-varying word sparse representations. It consists in learning an embedding matrix for the first time slice $t_0$, then updating it at each time step $t$ using the matrix at $t-1$ as initialisation. This method is called incremental updating. Another broadly used method it to learn an embedding matrix for each time slice independently; due to the stochastic aspect of word embeddings, the vectorial space for each time slice is different, making them not directly comparable. Thus, authors perform an alignment of the embeddings spaces by optimising a geometric transformation~\cite{Hamilton2016,Dubossarsky,Szymanski, Kulkarni}).
\\

In the case of sparse data, in addition to the approximative aspect of the alignment that harms the robustness of the embeddings, these methods are sensitive to random noise, which is difficult to disambiguate from semantic drifts. Moreover, the second one require large amounts of data for each time step to prevent overfitting. 
\newcite{Tahmasebi2018ASO} shows that low-frequency words have a much lower temporal stability than high-frequency ones.
In~\cite{Tahmasebi2018SurveyOC}, the authors explain that usual methods for diachronic embeddings training such as the two previously presented are inefficient in the case of low-frequency words and hypothesise that a new set of methods, pooled under the name of \emph{dynamic} models, may be more adapted. These models use probabilistic models to learn time-varying word embeddings while controlling the drift of the word vectors using a Gaussian diffusion process. \newcite{bamler17a} uses Bayesian word embeddings, which makes the algorithm more robust to noise in the case of sparse data; while \newcite{rudolph2018dynamic} relies on a Bernoulli distribution to learn the dynamic embeddings jointly across all time slices, making the most of the full dataset.\\


Outside of the framework of diachrony, several attempts aim at improving or adapting word embeddings to low-volume corpora in the literature. It can involve morphological information~\cite{luong2013better} derived from the character level~\cite{Santos14Character,Labeau15Non}, and often make use of external resources: semantic lexicon~\cite{faruqui2015retrofitting}, and pre-trained embeddings from larger corpora \cite{komiya2018investigating}.
However, to our knowledge, no work has attempted to apply similar solutions to the problem of sparse data in temporal corpora, even thought this situation has been faced by many authors, often in the case of short time steps for social media data \cite{Stewart2017VKontakte, bamler17a, Kulkarni}.


\section{Diachronic Models} \label{sec:model}

This section briefly describes the three models under study: the Skip-Gram incremental updating algorithm from~\newcite{kim2014}, the dynamic filtering algorithm of~\newcite{bamler17a}, and the dynamic Bernoulli embeddings model from \newcite{rudolph2018dynamic}. We consider a corpus divided into $T$ time slices indiced by $t$. For each time step $t$, every word $i$ is associated with two vectors $u_{it}$ (word vector) and $v_{it}$ (context vector).

\subsection{Incremental Skip-Gram (ISG)}
This algorithm relies on the skip-gram model estimated with negative sampling (SGNS) method described in~\cite{mikolovW2V} and it can be summarised as follows.  
The probability of a word $i$ to appear in the context of a word $j$ is defined by $\sigma(u_{i,t}^T v_{j,t})$, with $\sigma$ being the sigmoid function.
Words $i$ and $j$ are represented by their embedding vectors  $u_{i,t}$ and $v_{j,t})$ at time $t$. The matrices $U_t$ and $V_t$ gathers all of them for the whole vocabulary. The context is made of a fixed number of surrounding words and each word in the context are considered as independent of each other. 

The negative sampling strategy associates to each observed word-context pair (the positive examples) $n^+_{ijt}$, a set of negative examples $n^-_{ijt}$.  The negative examples are sampled for a noise distribution following~\newcite{mikolovW2V}. 

Let $n^{+-}_t$ denote for the time step $t$, the union of positive and negative examples. The objective function can be defined as the following log-likelihood:
\begin{multline} \label{SGloglik}
    log~p(n^{+-}_t|U_t,V_t) = \mathcal{L}_{pos}(U_t,V_t) + \mathcal{L}_{neg}(U_t,V_t) \\
    = \sum_{i,j=1}^L (n_{ijt}^+ log~\sigma(u_{i,t}^T v_{j,t}) + n_{ijt}^- log ~\sigma(-u_{i,t}^T v_{j,t}))
\end{multline}

For the first time slice, the matrices $U_1$ and $V_1$ are initialised using a Gaussian random noise $\mathcal{N}(0,1)$ before being trained according to equation~\eqref{SGloglik}. Then, for each successive time slice, the embeddings are initialised with values of the previous time slice following the methodology of \cite{kim2014}. This way, the word vectors of each time step are all in the same vectorial space and directly comparable.

\subsection{Dynamic Filtering of Skip-Gram (DSG)}

This second method relies on the Bayesian extension of the SGNS model described by~\newcite{barkan2015}. 
The main idea is to share information from one time step to another, allowing the embeddings to drift under the control of a diffusion process.
A full description of this approach, denoted as the filtering model, can be found in \cite{bamler17a}.

In this model, the vectors $u_{i,t}$ and $v_{i,t}$ are considered as latent vectors. Under a Gaussian assumption, they are represented by their means $(\mu_{u_{i,t}}, \mu_{v_{i,t}})$ and variances $(\Sigma_{u_{i,t}}, \Sigma_{v_{i,t}})$. 
They are initialised for the first time slice with respectively a zero mean vector and a identity variance matrix.


The temporal drift from one time step to another follows a Gaussian diffusion process with zero mean and variance $D$. 
This variance is called the \textit{diffusion} constant and has to be tuned along with the other hyperparameters. Moreover, at each time step a second Gaussian prior with zero mean and variance $D_0$ is added, resulting in the following distributions over the embeddings matrices $U_t$:
\begin{align} \label{eq:bamlerDrift}
    p(U_{1}|U_0) &\sim  \mathcal{N}(0,D_0)\\\nonumber
    p(U_t|U_{t-1}) &\sim  \mathcal{N}(U_{t-1},D) ~ \mathcal{N}(0,D_0).
\end{align}
The same equation stands for $V_t$. Training this model requires to estimate 
the posterior distributions over $U_t$ and $V_t$ given $n^{+-}_t$. This (Bayesian) inference step is unfortunately untractable. In~\cite{bamler17a}, the authors propose to use variational inference~\cite{Jordan99anintroduction} in its online extension~\cite{Blei2017VariationalIA}.
The principle of variational inference is to approximate the posterior distribution with a simpler variational distribution $q_\lambda(U,V)$ ($\lambda$ gathers all the parameters of $q$). This variational posterior will be iteratively updated at each time step. The final objective function can be written as follows: 

\begin{align}
  \label{elbo}
    \mathcal{L}_t(\lambda) &= \mathbb{E}_{q_\lambda} [ log~p(n^{+-}_t|U_t,V_t)] \\\nonumber
    &  + \mathbb{E}_{q_\lambda} [ log~p(U_t,V_t)|n^{+-}_{1:t-1}] \\\nonumber
    &- \mathbb{E}_{q_\lambda} [ log~q_\lambda(U_t,V_t)].
\end{align}

This loss function is the sum of three terms: the log-likelihood (computed following equation~\eqref{SGloglik}), the log-prior (which enforces the smooth drift of embedding vectors, sharing information with the previous time step), and the entropy term (approximated as the sum of the variances of the embedding vectors).



\subsection{Dynamic Bernoulli Embeddings (DBE)}

The DBE models extends the \textit{Exponential Family Embeddings} (EFE)\cite{rudolph2016exponential}, a probabilistic generalisation of the \textit{Continuous Bag-of-Words} (CBOW) model of~\newcite{mikolovW2V}. The main idea is that the model predicts the central word vector conditionally to its context vector following a Bernoulli distribution. A detailed description of the model can be found in \cite{rudolph2018dynamic}.

Each word $i$ has $T$ different embeddings vectors $u_{it}$, but this time, the context vectors $v_{i}$ are assumed to be fixed across the whole corpus. The embedding vector $u_{it}$ drifts throughout time following a Gaussian random walk, very similarly to equation (\ref{eq:bamlerDrift}):
\begin{align}\label{eq:deriv}
    U_0,~ V &\sim \mathcal{N}(0,\lambda_0^{-1}I), \\\nonumber
     U_t &\sim \mathcal{N}(U_{t-1},\lambda^{-1}I).
\end{align}

The \textit{drift} hyper-parameter $\lambda$ controls the temporal evolution of $U_t$, and is shared across all time steps.
The training process, described more precisely by \newcite{rudolph2018dynamic}, relies on a variant of the negative sampling strategy described by \newcite{mikolovW2V}. The goal is to optimise the model  across all time steps jointly, by summing over $t$ the following loss function:
\begin{multline} \label{eq:bernoulli_drift}
    \mathcal{L}_t =\mathcal{L}_{pos}(U_t,V) + \mathcal{L}_{neg}(U_t,V) \\
    +  \mathcal{L}_{prior}(U_t,V).
\end{multline}

The two first terms are computed as in equation (\ref{SGloglik}). The third term is defined as:
\begin{multline} \label{eq:bernoulli_prior}
    \mathcal{L}_{prior}(U_t,V) = - \frac{\lambda_0}{2} \sum_{i=1}^L \| v_{i} \|^2- \frac{\lambda_0}{2} \sum_{i=1}^L \| u_{i,0} \|^2 \\
    - \frac{\lambda}{2} \sum_{i,t} \| u_{i,t} - u_{i,t-1} \|^2. 
\end{multline}
The role of  $\mathcal{L}_{prior}$ is twofold: it acts as a regularisation term on $V$ and $U_t$, and as a constraint on the drift  of $U_t$, preventing  it from going too far apart from  $U_{t-1}$

\section{Experimental Results} \label{sec:expe}

The goal of this study is to compare the behaviour of the three algorithms described in section \ref{sec:model} in case of low-volume corpora. 
We evaluate their predictive power on different volumes of data to compare the impact of two initialisation methods, and analyse the behaviour of the drift of the embeddings.

\subsection{Experimental Setup} \label{sec:data}



We use the \textit{New York Times Annotated Corpus} (NYT) \cite{nytLDC} \footnote{released by the Linguistic Data Consortium} containing around 1~855~000 articles ranging from January $1^{st}$ 1987 to June 19th 2007. We divide the corpus into $T=20$ yearly time steps (the incomplete last year is not used in the analysis) and held out 10~\% of each time step for validation and testing. Then, we sample several subsets of the corpus: 50 \%, 10\%, 5\% and 1\% of the training set. This way, we can compare the models on each subset to evaluate their ability to train a model in the case of low-volume corpora.


We remove stopwords and choose a vocabulary of $V=10k$ most frequent words. Indeed, a small vocabulary is more adequate for sparse data in a temporal analysis in order to avoid having time steps were some word does not appear at all.
The total number of words in the corpus after preprocessing is around 38.5 million. It amounts to around 200k words per time step in the 10~\% subset of the corpus, thus only 20k in the 1~\% subset.



To tune the hyperparameters, we use the log-likelihood of positive examples $\mathcal{L}_{pos}$ measured on the validation set. We train each model for 100 epochs, with a learning rate of $0.1$, using the Adam optimiser.
For the DSG model, we use a diffusion constant $D = 1$ and a prior variance $D_0 = 0.1$ for both corpora.
For the DBE model, we use  $\lambda = 1$ and $\lambda_0 = 0.01$.

We choose an embedding dimension $d=100$, as the experiments show that a small embedding dimension, as in \cite{Stewart2017VKontakte}, leads to smoother word drifts and makes the model less sensitive to noise when the data is scarce.

We use a context window of 4 words and a negative ratio of 1; we observed that having a higher number of negative samples artificially increased the held-out likelihood, but equalised the drifts of all the words in the corpus. Thus, in an extreme scarcity situation, each negative sample has a high weight during training: the number of negative samples has to be very carefully selected depending on the amount of data.

\subsection{Impact of Initialisation on Sparse Data}

\begin{figure*}
\begin{center} 
\includegraphics[width=1\textwidth]{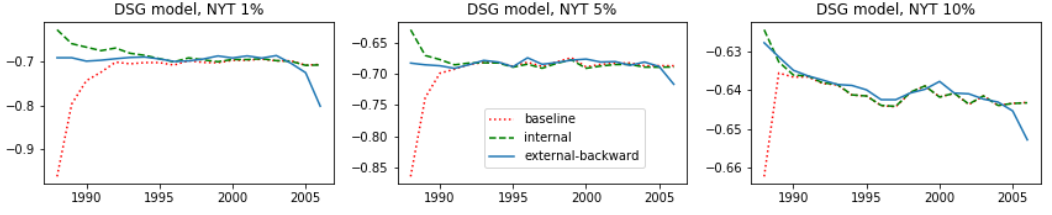}
\end{center} 
\caption{Log-likelihoods for the DSG model on three subsets of the corpus, comparing the baseline (random initialisation) with the two initialisation methods: \textit{internal} is the initialisation from the full dataset while \textit{external-backward} is the initialisation with the Wikipedia vectors, with training from most recent to oldest time step.}
\label{fig:heldout_init}
\end{figure*}

The embedding vectors of the ISG and DBE models are initialised using a Gaussian white noise, while the means and variances of DSG are initialised with null vectors and identity matrices respectively. However, a good initialisation can greatly improve the quality of embeddings, particularly in the case of scarce data.

We experiment the impact of two types of initialisation on the log-likelihood of positive examples on the test set.
\\

\textbf{Internal initialisation:}\\
We train each model in a static way on the full dataset. Then, we use the resulting vectors as initialisation for the first time step of the diachronic models. This methods is especially suitable for domain-specific corpora where no external comparable data is available.\\

\textbf{Backward external initialisation:}\\
We use a set of embeddings pre-trained on a much larger corpus for initialisation: The Wikipedia corpus (dump of August 2013) \cite{pretrainedEmbeddings} with vectors of size 100. These embeddings are representative of the use of words in 2013; and in general, large corpora exist almost exclusively for recent periods. Thus, we choose to use the pre-trained embeddings as initialisation for the \emph{last} time step (the most recent). Then, we update the embeddings incrementally from new to old (\textit{reverse incremental updating}).\\
This method would be particularly suitable for corpora with low volume in older time slices, as it is the case for most of the historical dataset in languages other than English.\\
For the DSG model, the pre-trained vectors are used as the mean parameter for each word. The variance parameter is fixed at 0.1. Experiments with a prior variance of 0.01 and 1 had a lowest log-likelihood on the validation set.\\

The log-likelihood curves in figure \ref{fig:heldout_init} show that the internal initialisation has a better impact on the likelihood at the beginning of the period, as it is closer to the data than the external initialisation. The positive impact of the backward external initialisation increases with the volume of data.

Overall, the mean log-likelihoods across all time steps (Table \ref{tab:loglik}) are  higher using the internal initialisation. We conjecture that internal initialisation is more profitable to the model when the period is short (here, two decades) with low variance. The backward external initialisation has very close scores to the internal one, and is more suitable for higher volume datasets with a longer period, as it gives higher benefit to the likelihood for bigger subsets.


\begin{table}[!ht]
\centering
\small
\begin{tabular}{|c|c|c|c|}
\hline
\textbf{\begin{tabular}[c]{@{}c@{}}Initialisation /\\ Model\end{tabular}} & \textbf{Random} & \textbf{Internal} & \textbf{\begin{tabular}[c]{@{}c@{}}Backward \\ external\end{tabular}} \\ \hline
\textbf{ISG}                                                              & -3.17          & -2.589            & -2.686                                                                \\ 
\textbf{DSG}                                                              & -0.749          & -0.686            & -0.695                                                                \\ 
\textbf{DBE}                                                              & -2.935          & -2.236            & -2.459                                                                \\ \hline
\end{tabular}
\caption{Log-likelihood on the 5\% subset of the NYT corpus for each model, with the three initialisation schemes.}
\label{tab:loglik}
\end{table}

\subsection{Visualising Word Drifts}

\begin{figure*}[ht!]
\begin{center} 
\includegraphics[width=1\textwidth]{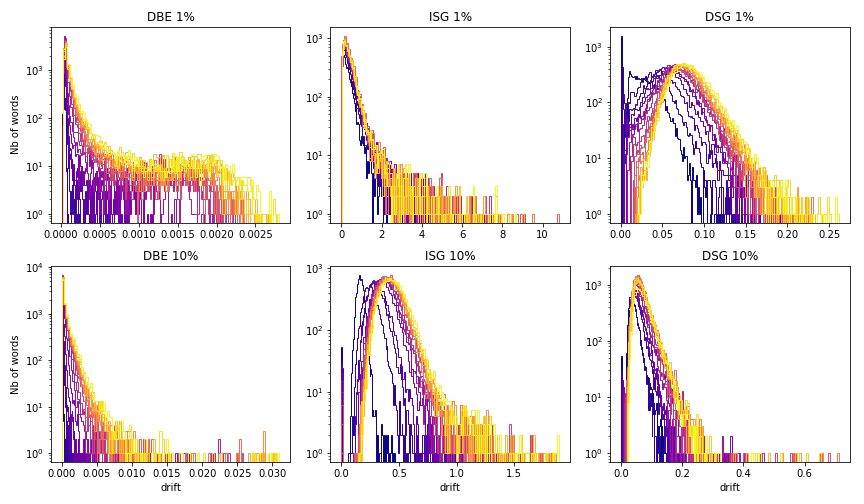}
\end{center} 
\caption{Histogram of word drift for each model on two subsets of the NYT corpus. The drifts are computed from $t_0=1987$ to each successive time step, and superposed on the histogram. The lightest colours indicate drifts calculated until the most recent time steps. The number of words are on logarithmic scale.}
\label{fig:hist}
\end{figure*}

A high log-likelihood performance does not necessarily imply that the drifts detected by the models are meaningful. In this section, we examine the distribution of word drifts outputted by each model with the internal initialisation. The computed drift is the L2-norm of the difference between the embeddings at $t_0$ and the embeddings at each $t$:
\begin{equation}
 \label{eq:drift}
    drift(U_t) =  \Big[\sum_{i=1}^L (u_{i,t}  - u_{i,t_0} )^2\Big]^{1/2}
\end{equation}

In the case of the DSG model were the words are represented as distributions, we compute the difference of the mean vectors.

We plot the superimposed histograms of \emph{successive} drifts (Figure \ref{fig:hist}) from $t_0=1987$ to each successive time step, for all studied models. For example, on the histograms, the lightest colour curve represents the drift between $t_0 = 1987$ and $t = 2006$ and the darkest one is the drift between $t_0 = 1987$ and $t = 1988$.

A first crucial property is the \emph{directed} aspect of the drifts: when the words progressively drift away from their initial representation in a directed fashion. On 10~\% of the dataset, the DBE model shows well this behaviour, with a very clear colour gradient. It is also the case for the other models on this subset. With 1~\% of the dataset on the contrary, the ISG model is unable to display a directed behaviour (no colour gradient), while the two other models do. This is justified by the use of the diffusion process to link the time steps in equations \ref{eq:bamlerDrift} and \ref{eq:bernoulli_drift}: it allows the DSG and DBE models to emphasise the directed fashion of drifts even in the situation of scarce data.

The second property to highlight is the capacity of the models to discriminate words that drift from words that stay stable. From the human point of view, a majority of words has a stable meaning \cite{Gulordava}; especially on a dataset covering only two decades like the NYT. The DBE model has a regularisation term (equation \ref{eq:bernoulli_prior}) to enforce this property, and a majority of words have a very low drift on the histogram. However, on 1~\% of the dataset, this model can not discriminate very high drifts from the rest. The ISG and DSG models have a different distribution shape, with the peak having a drift superior to zero. The only words that do not drift on their histograms are the one that are absent from a time step.


To conclude, both the DBE and DSG model are able to detect directed drifts even in the 1~\% subset of the NYT corpus, while the ISG can not. However, the drift distributions of the DBE and DSG models have a much shorter shorter tail on the 1~\% subset than on the 10~\% subset: they are not able to discriminate very high drifts from the rest of the words in extreme scarcity situation.

\subsection{Regularisation Attempt}

To tackle the weakness of the DBE and DSG models on the smallest subset, we attempt to regularise their loss in order to control the weight of the highest and lowest drifts. Our goal is to allow the model to: 
\begin{itemize}
    \item better discriminate very high drifts;
    \item be less sensitive to noise, giving lower weight to very low embedding drifts.
\end{itemize}{}

We test several possible regularisation terms to be added to the loss. The best result is obtained with the \emph{Hardshrink} activation function, which is defined this way: 
\begin{align}
    HardShrink(x)& = x, \textrm{ if } x > \beta\\\nonumber
     &= -x, \textrm{ if } x < -\beta\\\nonumber
      &= 0, \textrm{ otherwise}
\end{align}
For the DSG and DBE models, we add to the loss the following regularisation term, amounting to a thresholding function applied to the drift:
\begin{equation}
        \textrm{reg}_{\beta} = \alpha * HardShrink(\textrm{drift}(U_t),\beta)
\end{equation}
Where $\alpha$ is the regularisation constant to be tuned, $\beta$ is the threshold of the hardshrink function, and the drift is computed according to equation \ref{eq:drift}.
The regularisation term is minimised. The activation function acts as a threshold to limit the amount of words having an important drift. We choose $\beta$ as the mean drift for both models.\\

\begin{figure}[ht!]
\begin{center} 
\includegraphics[width=0.48\textwidth]{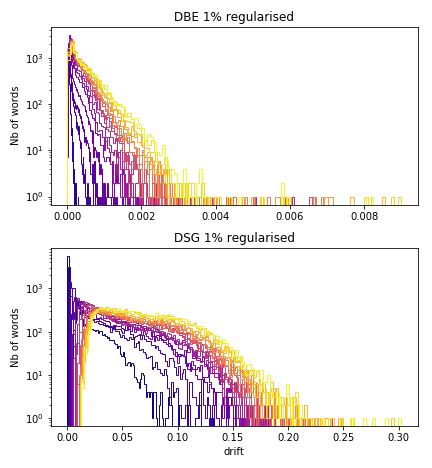}
\end{center} 
\caption{Histogram of word drift for the DBE and DSG regularised models on the 1~\% subset.}
\label{fig:freq_reg}
\end{figure}

For both DSG and DBE, the right tail of the distribution of the drifts with regularisation (Figure \ref{fig:freq_reg}) is much longer than in the original model (Figure \ref{fig:hist}). Moreover, in the case of the DSG model, more words have a drift very close to zero. 

To conclude, the regularised DSG model considers more words as temporally stable. Furthermore, regularising the loss of the dynamic models allows to better discriminate extreme word embedding drifts for very small corpora.

\section{Summary \& Future Work}

To summarise, we reviewed three algorithms for time-varying word embeddings: the incremental updating of a skip-gram with negative sampling (SGNS) from \newcite{kim2014} (ISG), the dynamic filtering applied to a Bayesian SGNS from \newcite{bamler17a} (DSG), and the dynamic Bernoulli embeddings model from \newcite{rudolph2018dynamic} (DSG), a probabilistic version of the CBOW. 

We proposed two initialisation schemes: the internal initialisation, more suited for low volume of data, and the backward external initialisation, more suited for higher volumes and long periods of temporal study.
Then, we compared the distributions of the drifts of the models. We conclude that even in extreme scarcity situations, the DBE and DSG models can highlight directed drifts while the ISG model is too sensitive to noise. Moreover, the DBE model is the best at keeping a majority of the words stable. This property, as long as the ability to detect directed drift, are two important properties of a diachronic model.
However, both have low ability to discriminate the highest drifts on a very small dataset. Thus, we added a regularisation term to their loss using the \emph{Hardshrink} activation function, successfully getting longer distribution tails for the drifts. \\


An important future work is the multi-sense aspect of words. Polysemy is a crucial topic when dealing with diachronic word embeddings, as the change in usage of a word can reflect various changes in its meaning. However, the more different senses are taken into account, the more data is needed to tackle it.  An evolution of the DSG model presented in this paper to adapt to this problem would be to represent words while taking into account the context of each occurrence of a word to disambiguate its meaning. \newcite{brazinskas} propose such model in a static fashion, where word vectors depends on the context and are drawn at token level from a word-specific prior distribution. The framework is similar to the Bayesian skip-gram model from \newcite{barkan2015} used in the DSG model; but the goal is to predict a distribution of meanings given a context for each word occurrence. We are working on adapting this model to a dynamic framework.



\bibliography{acl2019}
\bibliographystyle{acl_natbib}




\end{document}